%% file: root.tex
\documentclass[letterpaper, 10 pt, conference]{ieeeconf}  

\IEEEoverridecommandlockouts                              

\usepackage{epsfig} 
\usepackage{times} 
\usepackage{amsmath} 
\usepackage{amssymb}  
\usepackage{color}
\usepackage{xcolor}
\usepackage{preambles}
\usepackage{amsmath}
\usepackage{graphicx}
\usepackage{url} 
\usepackage[format=plain,labelsep=period]{caption}
\usepackage{wrapfig}
\usepackage[list=on,listformat=simple]{subcaption}
\usepackage{hyperref}
\usepackage{url}
\usepackage{makecell}
\usepackage{tabularray}
\usepackage{pbox}
\usepackage{caption}
\usepackage{float}
\floatstyle{plaintop}
\restylefloat{table}
\usepackage{subcaption}
\usepackage{mathtools}
\hypersetup{colorlinks = true, citecolor=black}
    
\usepackage{flushend}

\title{\LARGE \bf
Unified Path and Gait Planning for Safe Bipedal Robot Navigation}
\author{Chengyang Peng$^{1}$, Victor Paredes$^{1}$, and Ayonga Hereid$^{1}$
\thanks{*This work was supported in part by the National Science Foundation under grant FRR-21441568. }%
\thanks{$^{1}$Mechanical and Aerospace Engineering, Ohio State University, Columbus, OH, USA. {\tt\footnotesize (peng.947, paredescauna.1, hereid.1)@osu.edu.}}%
}

\usepackage[capitalise]{cleveref}

\usepackage[noadjust]{cite}

\usepackage{algorithm2e,algorithmic}

\begin{document}

\maketitle

\begin{abstract}

Safe path and gait planning are essential for bipedal robots to navigate complex real-world environments. The prevailing approaches often plan the path and gait separately in a hierarchical fashion, potentially resulting in unsafe movements due to neglecting the physical constraints of walking robots. A safety-critical path must not only avoid obstacles but also ensure that the robot's gaits are subject to its dynamic and kinematic constraints.
This work presents a novel approach that unifies path planning and gait planning via a Model Predictive Control (MPC) using the Linear Inverted Pendulum (LIP) model representing bipedal locomotion. This approach considers environmental constraints, such as obstacles, and the robot's kinematics and dynamics constraints. By using discrete-time Control Barrier Functions for obstacle avoidance, our approach generates the next foot landing position, ensuring robust walking gaits and a safe navigation path within clustered environments. We validated our proposed approach in simulation using a Digit robot in 20 randomly created environments. The results demonstrate improved performance in terms of safety and robustness when compared to hierarchical path and gait planning frameworks.



\end{abstract}

\input{sections/introduction}

\input{sections/method}
\input{sections/simulation_and_test}

\input{sections/conclusions}
\bibliographystyle{IEEEtran}
\bibliography{references}

\end{document}

%% file: sections/introduction.tex
\section{Introduction}

Bipedal robotics has always been a fascinating and challenging topic in robotics research. However, associated with its promising applications, bipedal robots exhibit complex and high-dimensional models that are particularly hard to use for planning purposes. A way to deal with complex models is to use a template-based model that can be used for gait planning, such as~\cite{castillo2023tempalte,paredes2022resolved}. 
Moreover, given the growing prevalence of robots operating in safety-critical environments, such as around people or crowded spaces, ensuring the safety of robot motion is becoming increasingly crucial for deploying these intelligent machines.

Traditional path planning algorithms, such as Dijkstra, A*, and RRT, primarily generate collision-free paths from the start to the goal point geometrically~\cite{sleumer1999exact,gasparetto2015path}. When implementing the resulting path on an actual robot, it is often necessary to utilize the path-following process, typically employing a closed-loop controller driven by the error between the reference and the actual robot paths~\cite{gasparetto2015path,normey2001mobile,xue2018solving}. 
In addition, these approaches generate paths without considering the robot's dynamics and kinematics constraints leading to infeasible or unsafe paths for robots, particularly bipedal robots, to follow.
Some planning algorithms integrate steering controllers and path planning, such as Kinodynamic RRT*, LQR-RRT*, and their variants~\cite{perez2012lqr,goretkin2013optimal,schouwenaars2004receding,huang2022informable}, to accommodate the dynamical constraints of mobile robots. Simplified dynamical models, such as differential-drive vehicles, are often considered for synthesizing dynamically consistent paths~\cite{papadopoulos2007differential,ji2016path}. While these algorithms can generate relatively smooth trajectories for mobile robots compared to geometric path-planning methods, they do not provide guaranteed control solutions applicable to bipedal robots due to the non-smooth and unstable nature of bipedal locomotion. Guiding a bipedal robot to follow these resultant paths still requires a separate path-following controller, either via way-point tracking or velocity tracking~\cite{peng2023safe}.

\begin{figure}
\centering
\vspace{2mm}
    \includegraphics[trim={2cm 0cm 2cm 0cm},clip,width=1\columnwidth]{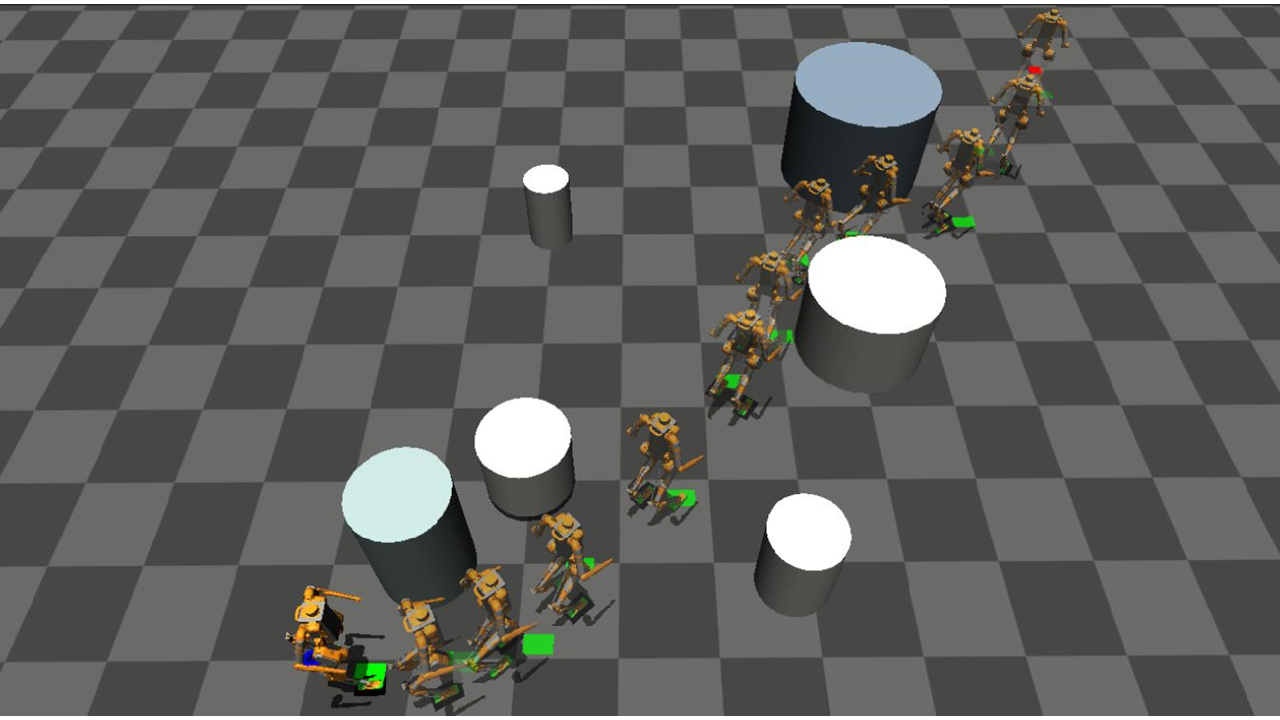}
\caption{\small{Safe navigation planning is tested in MuJoCo with the robot Digit, which generates stable walking gaits that avoid obstacles.}
} 
\label{fig:simu_nav_1}
\vspace{-3mm}
\end{figure}

Utilizing the full-order robot model for long-range path planning in bipedal locomotion poses significant challenges, primarily due to its demanding computational requirements. These challenges arise from the robot's high degrees of freedom, combined with its intricate and inherently unstable dynamics. Moreover, the system's sensitivity to uncertainties further complicates the complexity of real-time motion planning processes.
To mitigate these challenges, researchers often turn to simplified template models such as the Linear Inverted Pendulum Model (LIP)~\cite{kajita20013d,teng2021toward,gibson2022terrain}. By employing such reduced-dimensional representations, LIP-based approaches utilize preview-based optimization algorithms to determine the next stepping position of the swing foot. The primary objective is to stabilize the robot's center of mass (CoM) while maintaining alignment with the desired velocity trajectory~\cite{gibson2022terrain,teng2021toward,hsu2015control, liu2023realtime}. However, it's important to note that conventional LIP-based approaches typically do not account for the robot's turning. Consequently, the process of turning remains entirely independent of LIP-based gait planning approaches.

In this work, we present a unified path and gait planning framework for bipedal robots that generates collision-free, stable walking gaits. We formulate a multi-step preview plan based on the discrete-time step-to-step LIP model using a Model Predictive Control (MPC) formulation. This work is significant for two primary reasons: First, the MPC framework enables the incorporation of essential kinematic and dynamic constraints relevant to bipedal locomotion. These constraints include limitations on maximum forward and lateral speeds, rates of turning, the extremities of leg separation, and the interaction between walking speed and turning dynamics. Second, we integrate discrete-time Control Barrier Functions (D-CBFs) within our framework. This inclusion is pivotal for ensuring the generation of safe (i.e., collision-free) pathways in clustered environments. 
Control Barrier Functions (CBF) have been successfully applied in controlling legged robots~\cite{hsu2015control, ames2016control,ames2019control} and are now commonly used for ensuring safety in path planning~\cite{yang2019sampling,manjunath2021safe,ahmad2022adaptive,peng2023safe}. Specifically for bipedal robots, Liu et al. created an obstacle avoidance planner that combines Control Lyapunov Functions (CLF) with CBF constraints in a quadratic programming (QP) setup~\cite{liu2023realtime}. Agrawal et al. extended this idea to a discrete-time step planner with the Hybrid Zero Dynamics (HZD) based locomotion controller~\cite{agrawal2017discrete}. To accommodate long-range path planning, Teng et al. proposed a safety-critical multi-step planner based on the Linear Inverted Pendulum (LIP) model subject to Discrete-time CBF safety constraints~\cite{teng2021toward}. However, this work did not explicitly consider the heading angle and robot steering in the path planning process.

The rest of the paper is organized as follows: Section II reviews the background of the LIP model and introduces the new expression of the LIP model with heading angle state.
In Section III, we present the design process of the LIP-MPC by integrating the LIP model, MPC, and its relevant constraints.
Section IV showcases the utilization of LIP-MPC in simulation to implement the navigation of a bipedal robot, Digit. Through these simulations, we highlight the effectiveness and enhanced performance achieved by our LIP-MPC approach.
Finally, Section V briefly discusses the proposed work, its limitations, and some future directions.

%% file: sections/method.tex
\section{3D Linear Inverted Pendulum Model with Heading Angle}
The full-order dynamics of a bipedal robot are characterized by high dimensionality and non-linearity, rendering them impractical for path planning purposes.
By employing a template model, we can simplify the robot's dynamics into a lower-dimensional representation, avoiding the computational complexity of the full dynamics in planning. In our approach, we utilize the Linear Inverted Pendulum (LIP) model~\cite{kajita20013d}, anchored in a globally fixed frame to facilitate long-range path planning.

\begin{figure}
\centering
    \includegraphics[trim={2.3cm 0cm 3cm 2cm},clip,width=0.8\columnwidth]{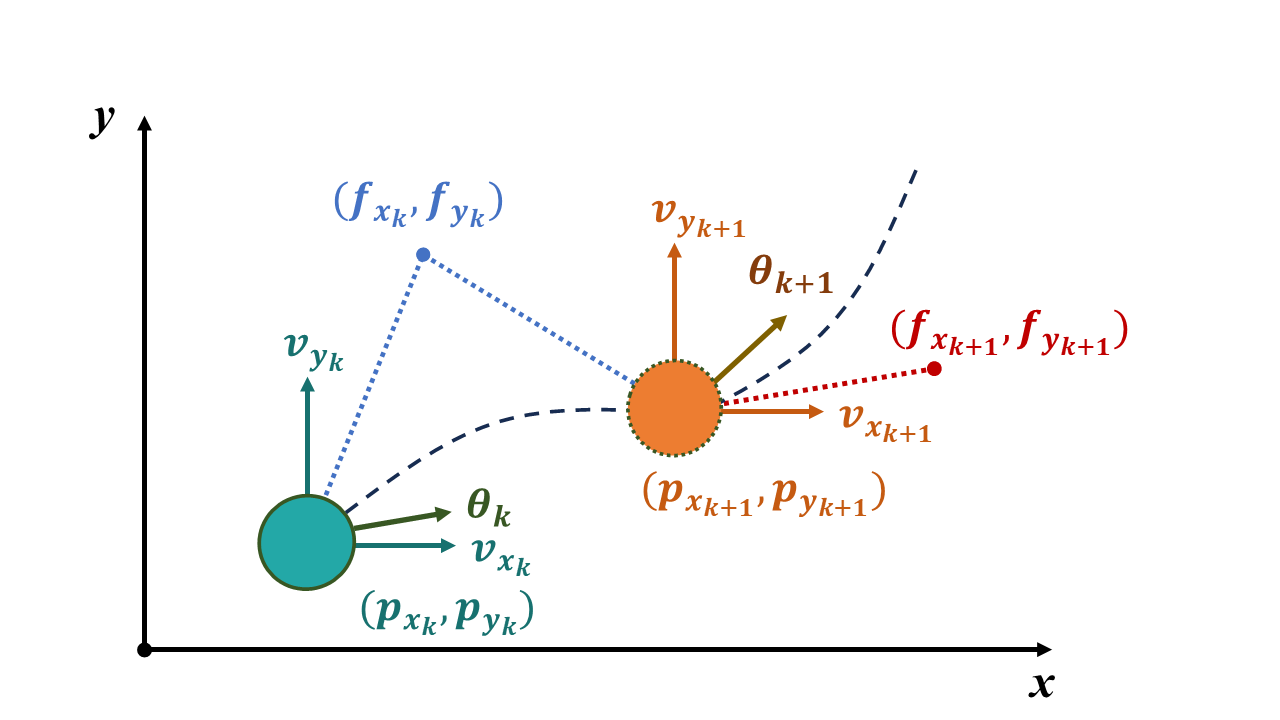}
\caption{The state transition of the Linear Inverted Pendulum model in the global frame. The state  $\mathbf{x}_{k}$ in step $k$ (shown in cyan) evolves to the state in the next step $\mathbf{x}_{k+1}$ (shown in orange) when the stance foot locates at point $(f_{x_k}, f_{y_k})$ and with a turning rate of $\omega_k$.}
\label{fig:LIP_new}
\vspace{-1mm}
\end{figure}

By maintaining a constant height of the Center of Mass (CoM), denoted as $H$, the motion of the robot along the $x$-axis can be effectively approximated by the linear dynamics of the LIP model, given by:
\begin{align}
    \begin{bmatrix} \dot{p}_{x} \\ \dot{v}_{x}\end{bmatrix} = 
    \begin{bmatrix} v_{x} \\ \frac{g}{H}(p_x-f_x) \end{bmatrix},
    \label{eq:lip-dynamics}
\end{align}
where $p_x$ and $v_x$ are the CoM position and velocity in $x$-axis,  $f_x$ is the stance foot position, and $g$ is the gravitational acceleration. We further assume that each walking step takes a fixed time duration $T$. Note that all positions are defined in the fixed global coordinate frame so that there is no need to reset the coordinate before and after a foot impact occurs.
Assuming that the position of the stance foot in a walking step $k$ is $f_{x_k}$, then the step-to-step discrete dynamics can be determined by: 
\begin{align}
\label{eq:x_dir_dynamics}
    \begin{bmatrix} p_{x_{k+1}} \\ v_{x_{k+1}} \end{bmatrix} =  \mathbf{A_d}
    \begin{bmatrix} p_{x_k} \\ v_{x_k} \end{bmatrix}
    + \mathbf{B_d}
    f_{x_k},
\end{align}
with: 
\begin{equation}
\label{eq:matrix_small}
\begin{aligned}
    \mathbf{A_d} &\coloneqq \left[ {\begin{array}{cc}
    \cosh(\beta T) & \frac{\sinh(\beta T)}{\beta}\\
   \beta\sinh(\beta T)  & \cosh(\beta T)
    \end{array} } \right],\\
    \mathbf{B_d} &\coloneqq \begin{bmatrix} 1-\cosh(\beta T) \\-\beta\sinh(\beta T) \end{bmatrix},
\end{aligned}
\end{equation}
where $\beta = \sqrt{g/H}$, and $p_{x_k}$ and $v_{x_k}$ represent the CoM position and velocity at the beginning of $k$-th step.

For 3D bipedal locomotion, the motion in the y-axis direction is similar to the x-axis, and therefore, the LIP model can describe it in the same fashion. In addition, we introduced a heading angle state $\theta$ and turning control $\omega$ to consider robot turning motion.
Let $\mathbf{x} \coloneqq [p_x, v_x, p_y, v_y, \theta]^T\in\mathcal{X}\subset\mathbb{R}^5$ and the control variable $\mathbf{u} \coloneqq [f_x, f_y, \omega]^T\in\mathcal{U}\subset\mathbb{R}^3$, in which the subscript $x$ and $y$ represent variables in $x$-axis and $y$-axis in the global coordinate frame. Thus, the step-to-step dynamics of the modified 3D-LIP model can be written as:
\begin{align}
\label{eq:system_dynamics}
    \mathbf{x}_{k+1} =  \mathbf{A}
    \mathbf{x}_k
    + \mathbf{B}\mathbf{u}_k,
\end{align}
with: 
\begin{equation}
\label{eq:system_matrix}
\begin{aligned}
    \mathbf{A} &\coloneqq \begin{bmatrix}
    \mathbf{A_d} & \mathbf{0} & \mathbf{0}\\
    \mathbf{0}& \mathbf{A_d} & \mathbf{0}\\
    \mathbf{0} &  \mathbf{0} & 1
    \end{bmatrix}, \quad
    \mathbf{B} &\coloneqq \begin{bmatrix} 
    \mathbf{B_d} & \mathbf{0} & \mathbf{0}\\
    \mathbf{0} & \mathbf{B_d}  & \mathbf{0}\\
    0 & 0 & 1
    \end{bmatrix}.
\end{aligned}
\end{equation}
An illustration of the LIP states and controls, projected in the $x-y$ plane, is shown in \figref{fig:LIP_new}. 
Defining the LIP dynamics in the global coordinate allows us to conveniently calculate obstacle avoidance in the form of CBF constraints in the following section.


\section{Unified Path and Gait Planning via LIP-MPC}

This section introduces an MPC formulation that unifies path and gait planning subject to robot dynamics approximated by the LIP model. 
\subsection{LIP-based Model Predictive Control}
Model Predictive Control (MPC) utilizes the prediction of system states over a finite time horizon to optimize control actions to realize desired performances by minimizing a cost function while satisfying specific constraints.

\begin{figure}
\centering
    \includegraphics[trim={0cm 0cm 1cm 3cm},clip,width=0.95\columnwidth]{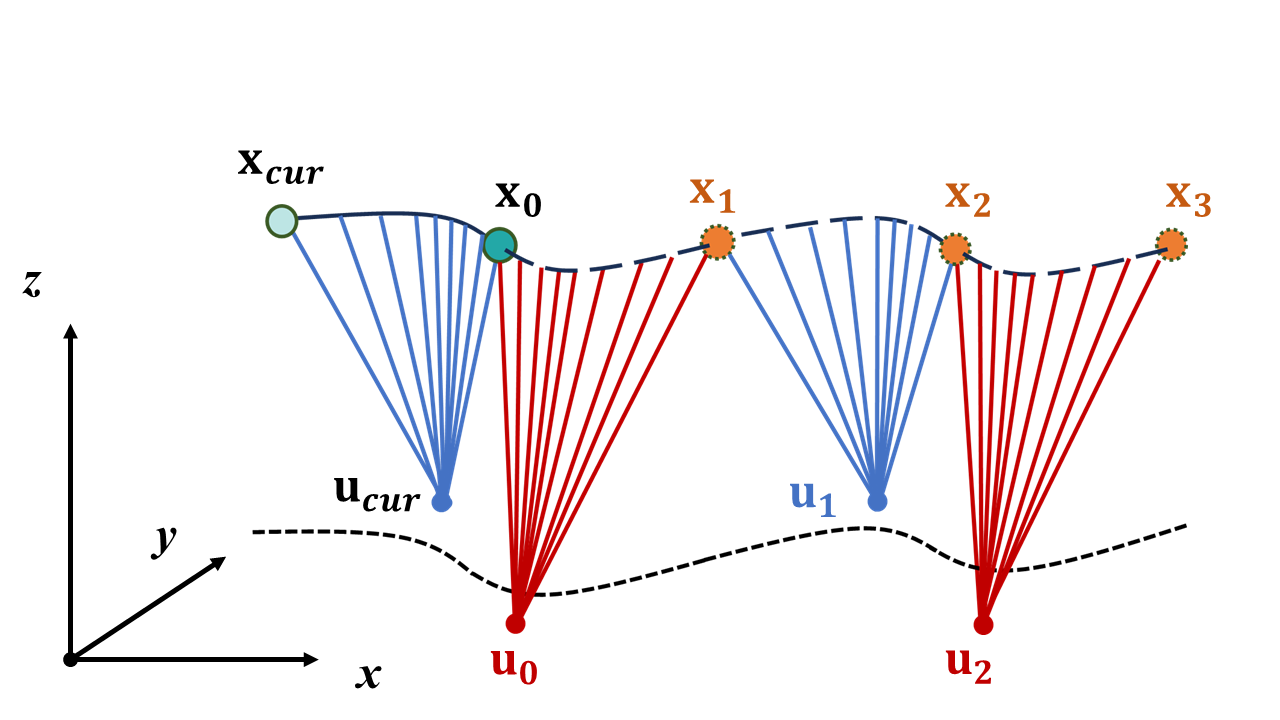}
\caption{An illustration of the LIP-MPC formulation when $N=3$. The planner first estimates the states at the end of the current step, $\mathbf{x}_0$, given the current CoM states, $\mathbf{x}_{cur}$, and the stance foot position, $\mathbf{u}_{cur}$. The LIP-MPC will determine the next three control actions (i.e., stepping positions and turning rates), as well as subsequent states for further evaluation.}
\label{fig:3_step_pre}
\vspace{-1mm}
\end{figure}

In our work, we use the step-to-step LIP dynamics in \eqref{eq:system_dynamics} to compute the optimal stepping positions of the next $N$ steps. Let $\mathbf{x}_{cur}$ be the LIP states at the current time, and $\mathbf{u}_{cur}$ is the current known stance foot position and turning rate. 
With the time remaining in the current step, we initially integrate the continuous LIP dynamics to ascertain the end-of-step LIP states, $\mathbf{x}_0$, which also serve as the starting point for the subsequent step.
This process sets the stage for formulating the LIP-MPC as follows:
\begin{align}
\label{eq:MPC_prob}
    J^* =& \min_{\mathbf{u}_{0:N-1}} \hspace{1em}  \sum_{k=1}^{N} q(\mathbf{x}_{k}) \\
    \st &\mathbf{x}_{k} \in \mathcal{X}, \text{ } k \in [1, N]\nonumber\\
    &\mathbf{u}_{k} \in \mathcal{U},\text{ } k \in [0, N-1]\nonumber\\
    &\mathbf{x}_{k+1} =  \mathbf{A}
    \mathbf{x}_{k}
    + \mathbf{B}\mathbf{u}_{k}, \text{ } k \in [0, N-1]\nonumber\\
    &\mathbf{c}_l \leq \mathbf{c}(\mathbf{x}_{k}, \mathbf{u}_{k}) \leq \mathbf{c}_u, \text{ } k \in [0, N-1]\nonumber
\end{align}
The decision variables in our approach are the next $N$ control actions with their corresponding states predicted using the 3D-LIP model, as shown in \figref{fig:3_step_pre} for $N=3$. Given the incorporation of nonlinear constraints within our framework, we use the IPOPT solver~\cite{wachter2006implementation} in this work.
The cost function $q(\mathbf{x}_k)$ is defined to evaluate a sequence of predicted states against the goal positions, aiming to minimize this cost to steer the robot towards its goal. In our design, we also consider the heading angle cost, defined as:
\begin{align}
\label{eq:theta_goal}
    \theta_{goal_k} = \arctan\left(\frac{p_{y_k}-y_{goal}}{p_{x_k}-x_{goal}}\right),
\end{align}
for each step $k$. We add the heading angle cost function to mitigate unnecessary adjustments and promote smoother navigation. Without this cost function, the planner tends to adjust heading angles frequently to realize maximum walking speeds. Thus, the cost function in each step is defined as:
\begin{align}
\label{eq:cost_finc_1_each}
    q(\mathbf{x}_{k})=&q_k\left(\left(p_{x_k}-x_{goal}\right)^2+\left(p_{y_k}-y_{goal}\right)^2\right) \nonumber \\
    & + r_k\left(\left(\theta_{k}-\theta_{goal_k}\right)^2\right),
\end{align}
where $q_k$ and $r_k$ are positive weights associated with position and orientation, respectively.

The function $\mathbf{c}(\mathbf{x}_{k}, \mathbf{u}_{k})$ represents the safety constraints for the bipedal robot. There are two types of safety constraints. The first is environmental safety constraints, primarily implemented by constructing Discrete Control Barrier Functions (DCBFs) to avoid a collision with the obstacles. Since planning a safe path does not guarantee safety when a bipedal robot walks along that path, we have added a second type of safety constraint. These constraints are constructed based on the kinematics of bipedal robots. 

\subsection{Environmental Safety Constraints}
Consider the discrete states of the robot $\mathbf{x}_k \in \mathcal{X}\subseteq \mathbb{R}^n$, and $\mathbf{u}_k \in \mathcal{U} \subseteq \mathbb{R}^m$ are the discrete control inputs. If there exists a continuous and differentiable function $h: \mathbb{R}^n \rightarrow \mathbb{R}$, the safety set $\mathcal{C}$ of the system may be defined as:
\begin{equation}
\label{eq:safety-set}
\begin{aligned}
    \mathcal{C} &= \{\mathbf{x}_k\in \mathcal{X} |h(\mathbf{x}_k) \geq 0\},\\
    \partial\mathcal{C} &= \{\mathbf{x}_k\in \mathcal{X} |h(\mathbf{x}_k) = 0\}.
\end{aligned}
\end{equation}
Then, $h(\cdot)$ is a discrete control barrier function (DCBF) if it follows the condition in the discrete-time domain for all $\mathbf{x}_k\in\mathcal{C}$\cite{agrawal2017discrete,zeng2021safety}:
\begin{equation}
\label{eq:cbf_off}
\begin{aligned}
    \exists \text{ } \mathbf{u}_k &\text{ s.t. } \bigtriangleup h(\mathbf{x}_k, \mathbf{u}_k) \geq -\gamma(h(\mathbf{x}_k)),
\end{aligned}
\end{equation}
where $\bigtriangleup h(\mathbf{x}_k, \mathbf{u}_k) \coloneqq h(\mathbf{x}_{k+1}) - h(\mathbf{x}_k)$, and $\gamma$ is a class $\kappa$ function that satisfies $0 < \gamma(h(x)) \leq h(x)$. In the discrete domain, $\gamma$ can be also a scalar that $0 < \gamma \leq 1$. So, a DCBF constraint can be written as:
\begin{equation}
\label{eq:cbf_simp}
\begin{aligned}
    \exists \text{ } \mathbf{u}_k &\text{ s.t. } h(\mathbf{x}_{k+1}) + (\gamma - 1)h(\mathbf{x}_k) \geq 0.
\end{aligned}
\end{equation}
In our case, we consider the obstacles in the environment to be circular or elliptical. Therefore the expression of the barrier functions $h(.)$ are:
\begin{equation*}
\label{eq:barriers}
\begin{aligned}
    \text{Circular: }h(\mathbf{x}_k) =& (p_{x_k}-x_c)^2+(p_{y_k}-y_c)^2-r^2\\
    \text{Elliptical: }h(\mathbf{x}_k) =& A(p_{x_k}-x_e)^2+B(p_{x_k}-x_e)(p_{y_k}-y_e)\\
    +& C(p_{y_k}-y_e)^2-D^2
\end{aligned}
\end{equation*}
Where $[x_c, y_c, r]$ represents the center and radius of the circular obstacles. For elliptical obstacles, $[x_e, y_e]$ represents the geometric center, and parameters $A, B, C, D$ determine the shape and size of the obstacles, which can be determined by ellipse width, length, and rotation angle.
The LIP-MPC generates a collision-free path for the Center of Mass (CoM) references in global coordinates. However, to maintain the robot's end-effectors, like feet and hands, at a safe distance from obstacles, we consider a safety margin by enlarging the original obstacle size by $0.4$m. This adjustment simplifies the enforcement of environmental safety and improve collision avoidance.

\subsection{Kinematics Safety Constraints}
Since our LIP-MPC implements path and gait planning simultaneously, we need to consider the safety of the planned path (avoiding obstacles) as well as the safety and feasibility of the robot's actions. 
A planned optimal path may require the robot to cross its legs, stretch them beyond limits, or be compelled to maintain maximum forward speed while making a large turn. While these situations may be acceptable at the mathematical level for the LIP model, they pose safety risks for the actual robot during walking. Such scenarios can easily lead to the robot falling or even being damaged. Hence, incorporating these kinematics constraints is essential for comprehensive safety and feasibility planning.



\newsec{Linear Velocity.} One key constraint we address involves the robot's linear velocities in its local coordinate, whose $x$-axis points along its heading angle as shown in \figref{fig:LIP_new}. Bipedal robots have the capability to move in all directions; however, their design allows for quicker movement longitudinally compared to laterally. Because the LIP model describes states in global coordinates, applying these velocity constraints requires converting them into the robot's local coordinate frame. Additionally, the sign of lateral velocity at the end of a step is determined by the stance foot—it's positive when the right foot is the stance foot and negative for the left foot as the stance.
The inclusion of the heading angle in the LIP states allows for stating the velocity constraints as follows. When the right foot is the stance foot, we have
\begin{align}
\label{eq:rigt_vel_cons}
    \begin{bmatrix} v_{x_\mathrm{min}}\\ v_{y_\mathrm{min}} \end{bmatrix} \leq  \begin{bmatrix} \cos\theta_k & \sin\theta_k \\ -\sin\theta_k &  \cos\theta_k\end{bmatrix}
    \begin{bmatrix} v_{x_k} \\ v_{y_k} \end{bmatrix}
    \leq \begin{bmatrix} v_{x_\mathrm{max}}\\ v_{y_\mathrm{max}} \end{bmatrix},
\end{align}
where $v_{x_{\mathrm{min}}}, v_{x_{\mathrm{max}}}, v_{y_{\mathrm{min}}}, v_{y_{\mathrm{max}}}$ are the lower bounds and upper bounds of the robot longitudinal velocity and lateral velocity, respectively. 
When the stance foot is the left foot, the constraints can be expressed as follows:
\begin{align}
\label{eq:left_vel_cons}
    \begin{bmatrix} v_{x_\mathrm{min}}\\ -v_{y_\mathrm{max}} \end{bmatrix} \leq  \begin{bmatrix} \cos\theta_k & \sin\theta_k \\ -\sin\theta_k &  \cos\theta_k\end{bmatrix}
    \begin{bmatrix} v_{x_k} \\ v_{y_k} \end{bmatrix}
    \leq \begin{bmatrix} v_{x_\mathrm{max}}\\ -v_{y_\mathrm{min}} \end{bmatrix}.
\end{align}
Imposing strict constraints on the sign of lateral velocity at the end of each step can implicitly prevent leg crossing during locomotion. This measure is crucial since crossing legs can lead to collisions between the two legs, significantly increasing the risk of the robot falling.

\newsec{Leg Reachability.} We impose a swing foot reachability constraint to prevent over-extensioning the leg and violating the physical limits of leg joints. In particular, we limit the horizontal distance from the robot's CoM position $(p_{x_k},p_{y_k})$ to the subsequent stepping position $(f_{x_k},f_{y_k})$, ensuring it remains within a defined threshold. This constraint is formulated as:
\begin{align}
\label{eq:leg_cons}
    0 < (p_{x_k}-f_{x_k})^2+(p_{y_k}-f_{y_k})^2
    \leq \mathcal{L}_\mathrm{max}^2,
\end{align}
where $\mathcal{L}_\mathrm{max}$ is the maximum reachable distance of the swing foot on the ground.

\newsec{Turning Rate.} Considering a turning rate $\omega$, we directly impose a turning constraint with maximal turning rate $\Omega_{max}$ to prevent the LIP model from generating an excessive turning motion:
\begin{align}
\label{eq:heading_cons}
    -\Omega_{\mathrm{max}} \leq \omega_k
    \leq \Omega_\mathrm{max}.
\end{align}

\newsec{Maneuverability Constraint.} Decelerating while turning is an important strategy that can improve safety during walking. This approach mirrors human walking behavior, where there's a documented linear relationship between longitudinal velocity and changes in the heading angle, as observed in previous studies~\cite{courtine2003human, choi2019muscle, lopez2019walking}. The maneuverability constraint couples the turning rate with the longitudinal velocity, and can be expressed as follows:
\begin{align}
\label{eq:hd_vel_cons}
     v_{x_\mathrm{min}} \leq \frac{\alpha}{\pi}|\omega_k|+ (v_{x_k}\cos\theta_k + v_{y_k}\sin\theta_k) \leq v_{x_\mathrm{max}},
\end{align}
    where $\alpha$ is a positive coefficient that balances the turning rate and longitudinal velocity. A higher value means a greater need for deceleration to achieve the same angular displacement during a turn. In our work, we empirically set $\alpha = 3.6$ for the Digit robot based on our observation.
We design these constraints in our LIP-MPC to achieve a specific behavior: when the robot walks straight ahead (with a turning rate of 0), it can attain its maximum longitudinal velocity, $v_{x_{max}}$. However, when the robot turns, either left or right, it needs to decrease the longitudinal velocity in response to turning motion. This adjustment will minimize the oscillation of the turning rate, facilitating smoother turns.

\begin{figure}
\centering
\vspace{2mm}
    \includegraphics[trim={2.4cm 0.5cm 2.4cm 2.0cm},clip,width=0.85\columnwidth]{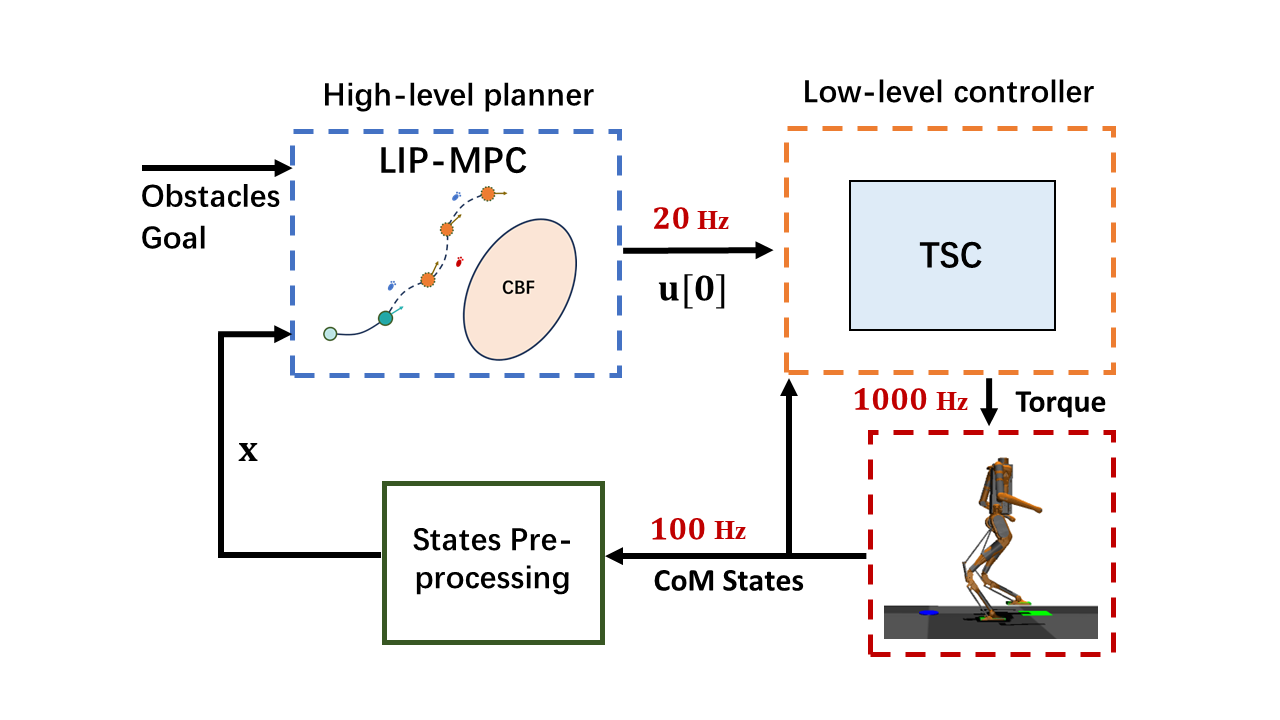}
\caption{The hierarchical framework for planning and control for safe bipedal navigation. The high-level planner unifies both path and gait planning, whereas the low-level task space controller tracks the target swing foot-stepping positions to follow the planned path. }
\label{fig:heriachical}
\end{figure}

\subsection{Hierarchical Locomotion Control Framework}

To implement safe navigation of the robot using the unified LIP-MPC planner, we implemented a hierarchical locomotion controller, as shown in \figref{fig:heriachical}. 

The high-level LIP-MPC planner takes environmental information and robot states as inputs, and determines the next foot-stepping position. In this work, we assume that the goal position remains constant and the geometric characteristics of static obstacles are predefined and known. The LIP-MPC is recalculated several times during a single walking step, leveraging real-time feedback on the robot's Center of Mass (CoM) states, resulting in an update rate of 20 Hz for the high-level planner. A state pre-processing component predicts the end-of-step states from the current robot states, factoring in the remaining time for the current step. This prediction acts as the initial state for the next cycle of the discrete-time LIP-MPC in \eqref{eq:MPC_prob}.
This iterative process enables the planner to dynamically adjust the positioning of the next swing foot, ensuring the robot maintains stability and avoids obstacles effectively over extended distances.

The low-level task space controller (TSC) runs at 1 kHz and determines the motor torques required for each joint to track the target swing foot positions, while maintaining a constant CoM height and ensuring the robot's torso remains upright. Further details on the low-level task space controller are available in our prior work~\cite{castillo2023tempalte}. 

%% file: sections/simulation_and_test.tex
\section{Simulation Results}

This section presents simulation results that demonstrate the effectiveness and improved performance of our proposed LIP-MPC in navigating clustered environments\footnote{A video showing all simulation results can be found in the \url{https://youtu.be/ES-Fmx3TOQs}.}.

\subsection{Simulation Setup}


In particular, we use the Agility Robotics' Digit humanoid as our testing platform in simulation.
To evaluate the efficacy and performance of our proposed method, we designed several test environments within the MuJoCo simulator, each featuring six to eight obstacles of varying sizes and shapes. For all test scenarios, the starting position was set at $[0,0]$ m, and the goal position was designated at $[10,10]$ m. Specifically, we maintained the robot's Center of Mass (CoM) height at a constant $H=1$ m, set the step duration to $T=0.4$ s, and defined the MPC prediction horizon as $
N=3$. \tabref{table:2} shows the chosen values of weights and limits we used throughout all tests in this paper. 

\begin{table}[h]
\vspace{2.5mm}
\centering
\begin{tabular}{l l  } \hline
Parameters & Value \\ \hline
$q_k$ & 1 \\
$r_k$ & 50 \\
$[v_{x_\mathrm{min}}, v_{x_\mathrm{max}}]$ & $[0.4, 0.8] $ m/s \\
$[v_{y_\mathrm{min}}, v_{y_\mathrm{max}}]$ & $[0.15, 0.35]$ m/s\\
$\mathcal{L}_{\mathrm{max}}$ & $0.3$m \\
$\Omega_{max}$ & $\frac{\pi}{16}$ rad/T \\
$\alpha$ & 3.6\\
$\gamma$ & $0.3$ \\
\hline
\end{tabular}
\caption{The value of each control parameter used in the simulation throughout this work.}
\label{table:2}
\vspace{-2mm}
\end{table}

\subsection{Improved Performance with Maneuverability Constraint and Heading Angle Cost}

One important contribution of this work is the incorporation of a heading angle cost and maneuverability constraints into the planning framework. These modifications facilitate smoother navigation and safer locomotion in clustered environments. To evaluate their impact, we conducted tests comparing the performance of the LIP-MPC with these modifications ({Modified LIP-MPC}) against the version without them ({Original LIP-MPC}). This comparison aims to clearly demonstrate the benefits these adjustments bring to the overall planning and execution of bipedal robot locomotion.


\begin{figure}
\centering
\vspace{2mm}
    \begin{subfigure}[b]{0.49\linewidth}
         \centering
        \includegraphics[trim={0.8cm 0.7cm 1cm 1.5cm},clip,width=\linewidth]{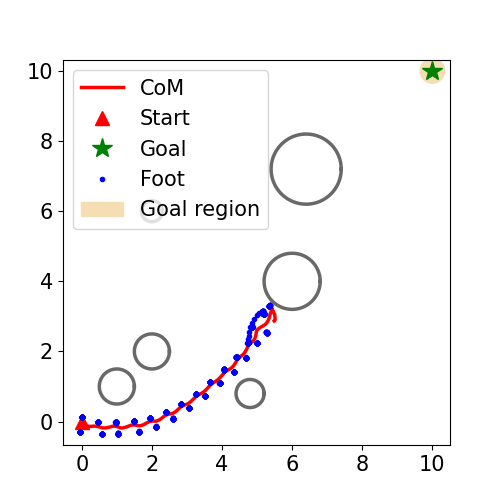}
     \end{subfigure}
     \begin{subfigure}[b]{0.49\linewidth}
         \centering
        \includegraphics[trim={7cm 0cm 7cm 0cm},clip,width=\columnwidth]{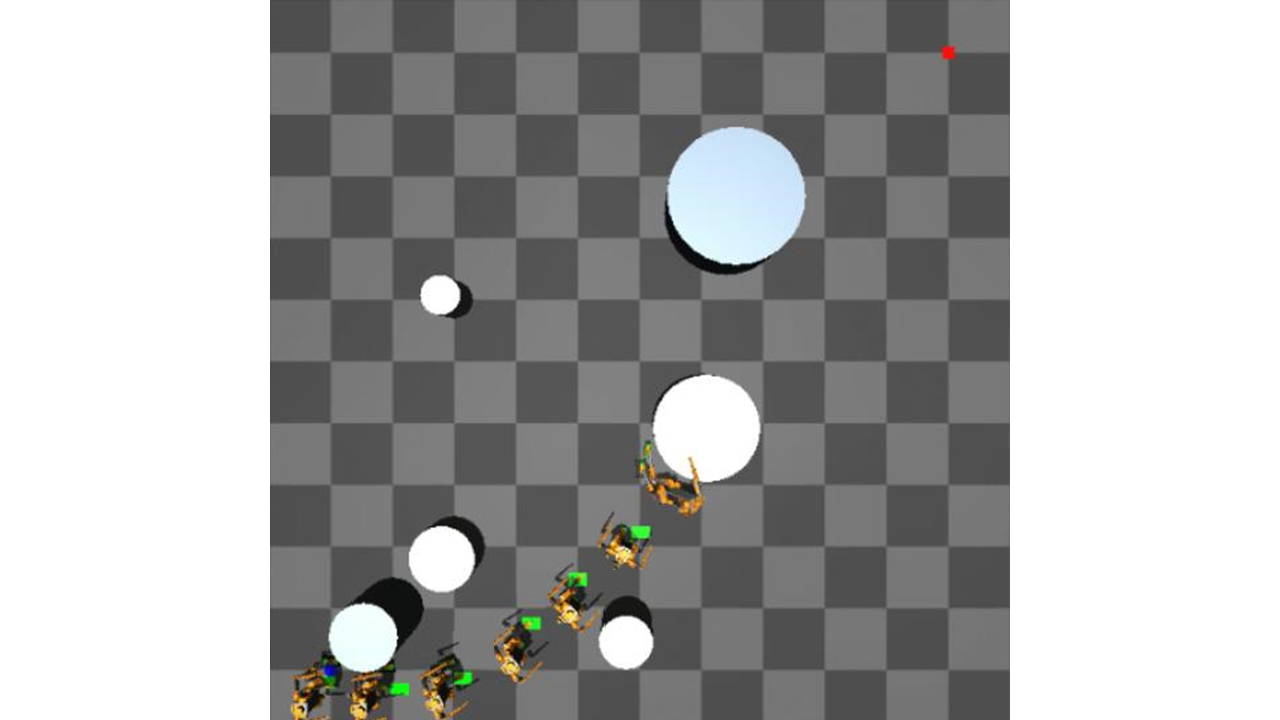}
     \end{subfigure}
\caption{Actual trajectories using the original LIP-MPC planner without considering heading angle cost and maneuverability constraints. Left shows the resultant CoM global position (m), and Right shows the robot colliding with obstacles and falling in simulation.}
\label{fig:orig_lip_res}
\end{figure}

\begin{figure}
\centering
\vspace{2mm}
    \begin{subfigure}[b]{0.49\linewidth}
         \centering
        \includegraphics[trim={0.8cm 0.7cm 1cm 1.5cm},clip,width=\linewidth]{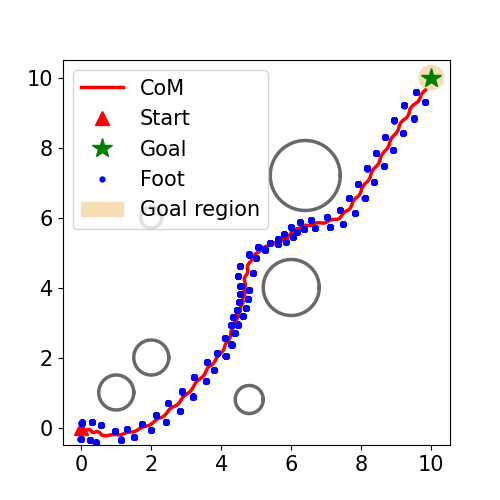}
         \caption{}
         \label{fig:modi_lip_path}
     \end{subfigure}
     \begin{subfigure}[b]{0.49\linewidth}
         \centering
        \includegraphics[trim={7cm 0cm 7cm 0cm},clip,width=\columnwidth]{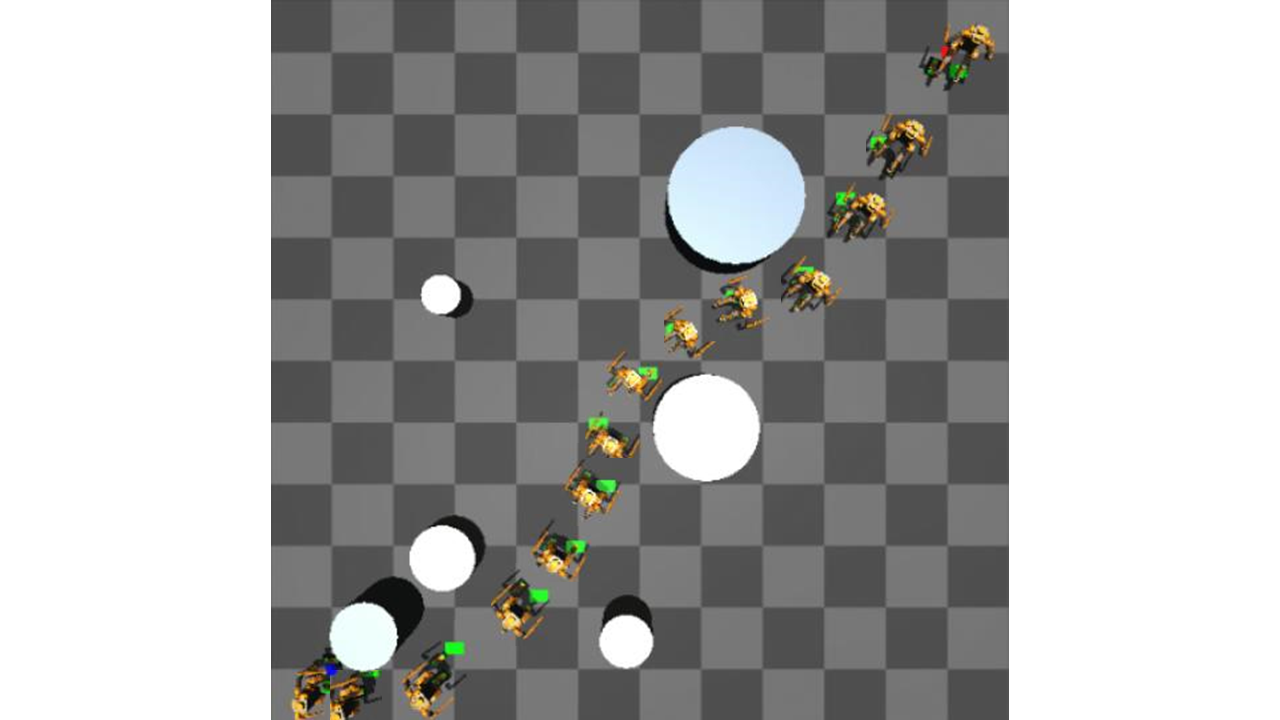}
        \caption{}
        \label{fig:modi_lip_sim}
     \end{subfigure}
\caption{Actual trajectories using the modified LIP-MPC planner adding angle cost and maneuverability constraints. Left shows the resultant CoM global position (m), and Right shows the robot successfully reaching the goal in simulation.}
\label{fig:modi_lip_res}
\end{figure}

\begin{figure}
\centering
\vspace{2mm}
    \begin{subfigure}[b]{0.49\linewidth}
         \centering
        \includegraphics[trim={7cm 0.1cm 8.6cm 2.1cm},clip,width=\linewidth]{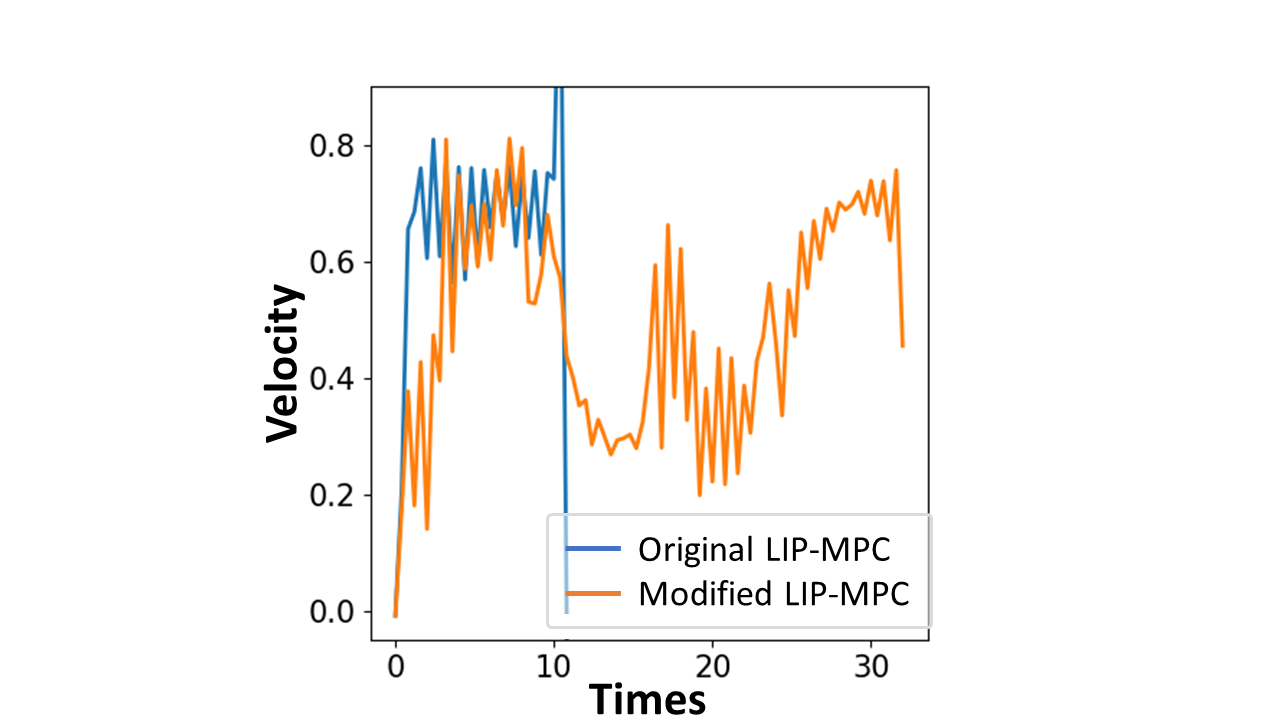}
     \end{subfigure}
     \begin{subfigure}[b]{0.49\linewidth}
         \centering
        \includegraphics[trim={7cm 0.1cm 8.6cm 2.1cm},clip,width=\linewidth]{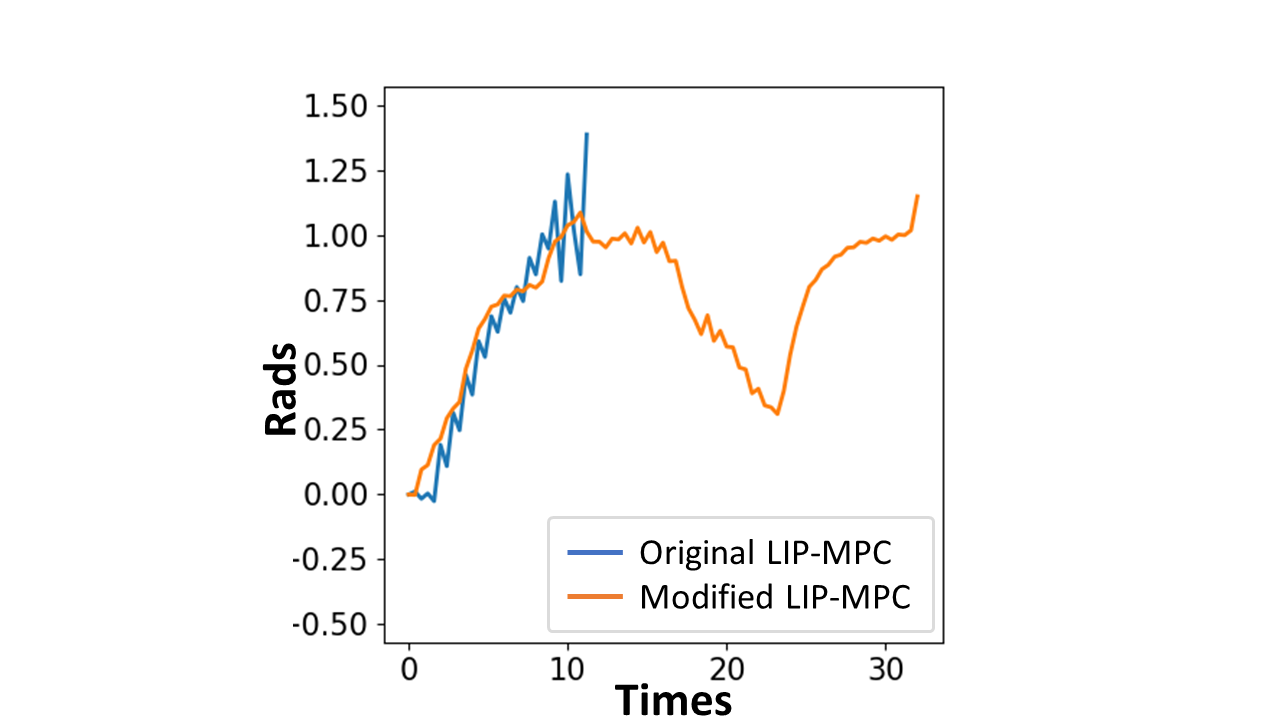}
     \end{subfigure}
\caption{States comparison of the original LIP-MPC and modified LIP-MPC. The left shows the longitudinal velocity (m/s) of two planners. The right shows the heading angle of two planners.}
\label{fig:vel-hd_comp}
\end{figure}

The comparative simulation results of the Original LIP-MPC and the Modified LIP-MPC are shown in \figref{fig:orig_lip_res} and \figref{fig:modi_lip_res}, respectively. The Original LIP-MPC encountered an obstacle collision leading to a fall, in contrast to the Modified LIP-MPC, which adeptly completed the navigation task. Further analysis of longitudinal velocity and heading angle changes throughout the trials in \figref{fig:vel-hd_comp} shows that the Original LIP-MPC tends to walk at the maximum longitudinal velocity regardless of the robot turning by frequently altering the robot's heading angle. This resulted in significant oscillation of the heading angle and a lack of deceleration during turns, which compromised the robot's stability, leading to a notable disparity between the actual CoM trajectory and the planned path. Consequently, the risk of the robot colliding with obstacles was noticeably increased. Moreover, such aggressive maneuvers could pose risks to the physical robot on hardware implementation. On the other hand, the Modified LIP-MPC planner adopted a more cautious approach by reducing speed during turns and allowing for smoother heading angle transitions. The planner only reaches the maximum velocity when walking in a straight line. These strategic modifications remarkably enhanced the safety and reliability of the LIP-MPC in navigating through obstacles.



\subsection{Improved Safety with LIP-MPC}

To show the enhancement in path and gait safety through the incorporation of robot dynamics approximated by the LIP model, we compare our method with a conventional hierarchical planning framework. This framework consists of two components: a high-level path planner that considers the robot as a differential drive and a mid-level gait planner that tracks desired walking velocities. For comparison purposes, we utilized a LIP-based gait planner, described in our previous work~\cite{gao2023time, paredes2022resolved}. The gait planner will track the desired velocity commands from the path planner to follow the planned path. 
The high-level differential-drive (DD) based path planner utilizes the same MPC formulation (referred to here as DD-MPC) with discrete-time Control Barrier Functions for obstacle avoidance described in the previous section. However, while DD-MPC can accommodate linear velocity and turning rate constraints, it lacks the capability to enforce additional kinematic constraints, such as leg reachability. Furthermore, due to the differential drive model's nonholonomic dynamics, DD-MPC only generates longitudinal velocity and turning rate, underscoring a limitation in its adaptability to complex locomotion tasks.







\begin{figure}
\centering
\vspace{2mm}
    \begin{subfigure}[b]{0.49\linewidth}
         \centering
            \includegraphics[trim={0.8cm 0.7cm 1cm 1.5cm},clip,width=\columnwidth]{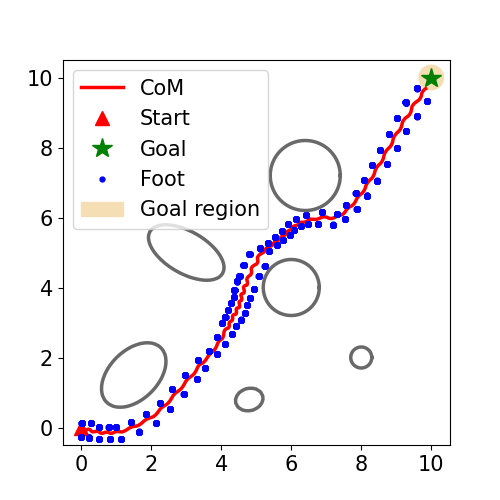}
     \end{subfigure}
     \begin{subfigure}[b]{0.49\linewidth}
         \centering
        \includegraphics[trim={7cm 0cm 7cm 0cm},clip,width=\columnwidth]{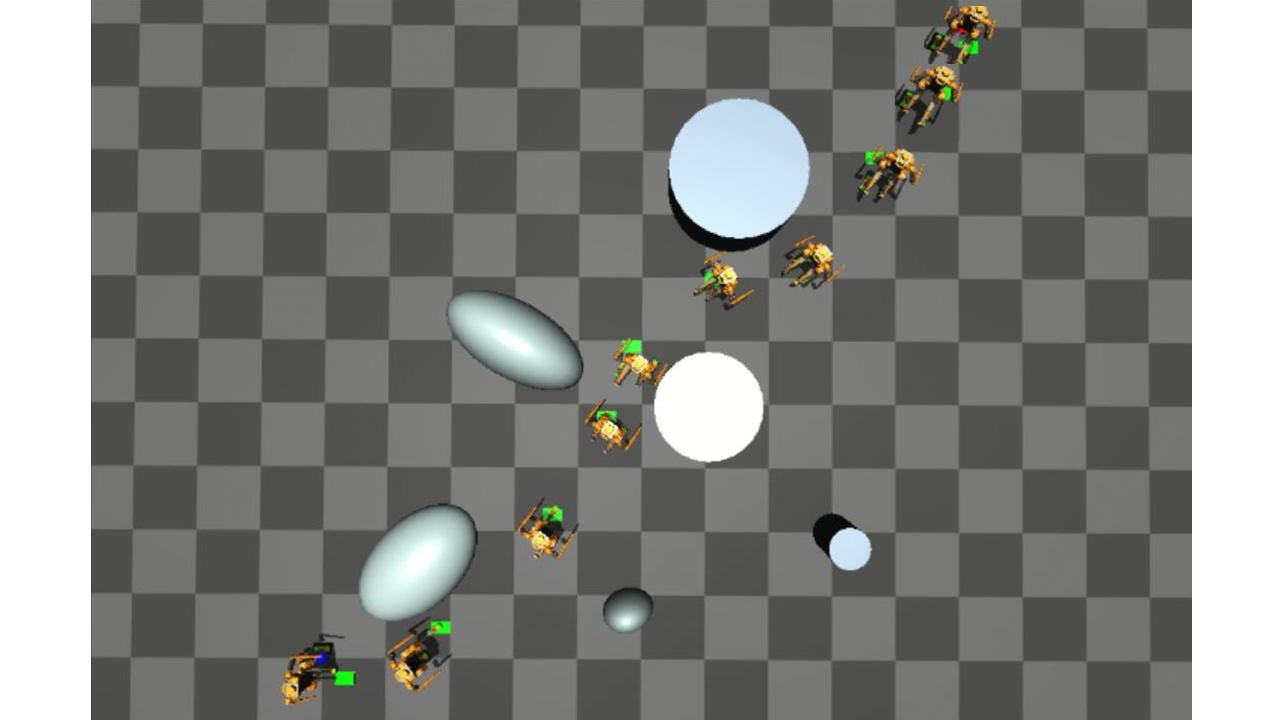}
     \end{subfigure}
\caption{Actual trajectories using the LIP-MPC planner. Left shows the resultant CoM global position (m), and Right shows the robot successfully reaching the goal in simulation.}
\label{fig:env_lip}
\end{figure}

\begin{figure}
\centering
\vspace{2mm}
    \begin{subfigure}[b]{0.49\linewidth}
         \centering
            \includegraphics[trim={0.8cm 0.5cm 1cm 1.5cm},clip,width=\columnwidth]{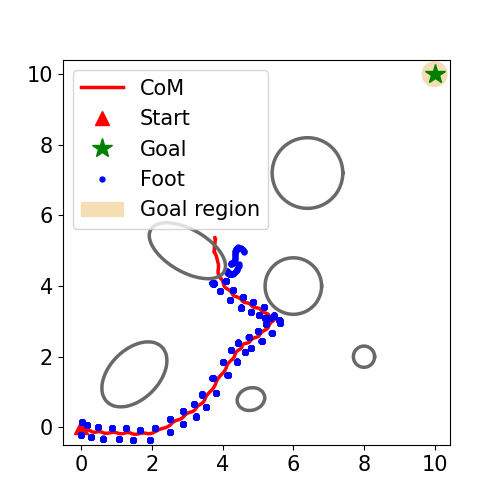}
     \end{subfigure}
     \begin{subfigure}[b]{0.48\linewidth}
         \centering
        \includegraphics[trim={7.2cm 0cm 7.5cm 0cm},clip,width=\columnwidth]{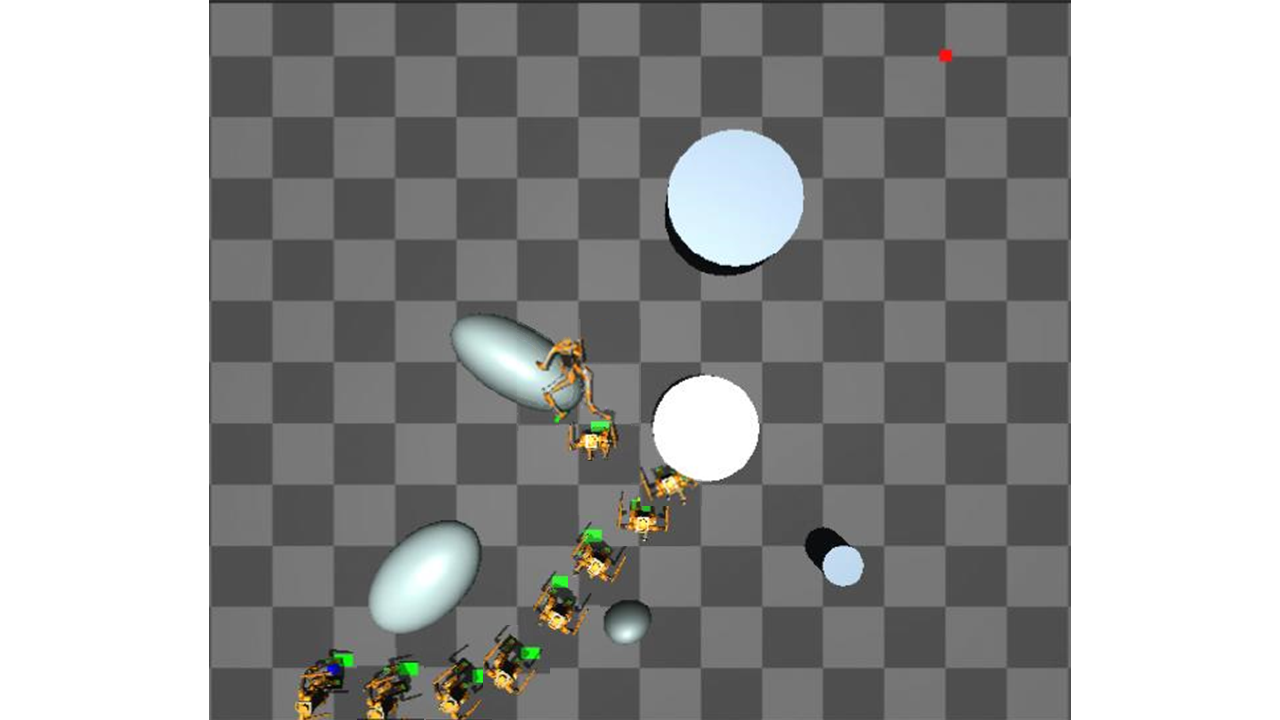}
     \end{subfigure}
\caption{Actual trajectories using the DD-MPC planner. Left shows the resultant CoM global position (m), and Right shows the robot collides with an obstacle and falls in simulation.}
\label{fig:env_dd}
\end{figure}

To evaluate the effectiveness of incorporating robot dynamics through the LIP model for safer path and gait planning, we conducted simulations in environments with six obstacles (three circular and three elliptical) to compare the performance of our LIP-MPC planner with the DD-MPC planner.
The simulation setup aimed to navigate the robot from an identical start to an end point across both planners. The simulation results are shown in \figref{fig:env_lip} and \figref{fig:env_dd}, respectively. \figref{fig:env_lip} shows that the robot successfully navigated around all obstacles to reach the goal without colliding and falling. On the other hand, the DD-MPC planner resulted in a collision with an obstacle, eventually causing the robot to fall. 

\begin{figure}
\centering
\vspace{2mm}
    \begin{subfigure}[b]{0.49\linewidth}
         \centering
            \includegraphics[trim={8.2cm 1cm 8.2cm 1.5cm},clip,width=\columnwidth]{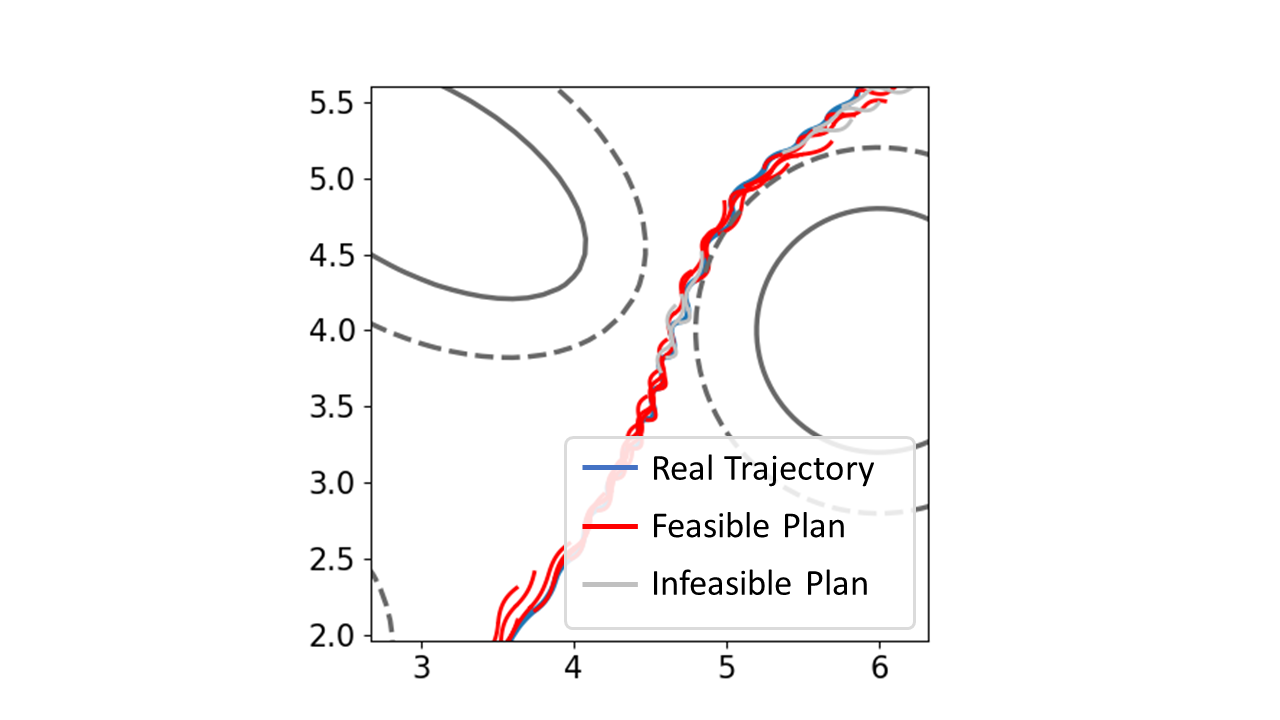}
         \caption{\small{LIP-MPC}}
         \label{fig:cir_det_LIP}
     \end{subfigure}
     \begin{subfigure}[b]{0.49\linewidth}
         \centering
        \includegraphics[trim={8.2cm 1cm 8.2cm 1.5cm},clip,width=\columnwidth]{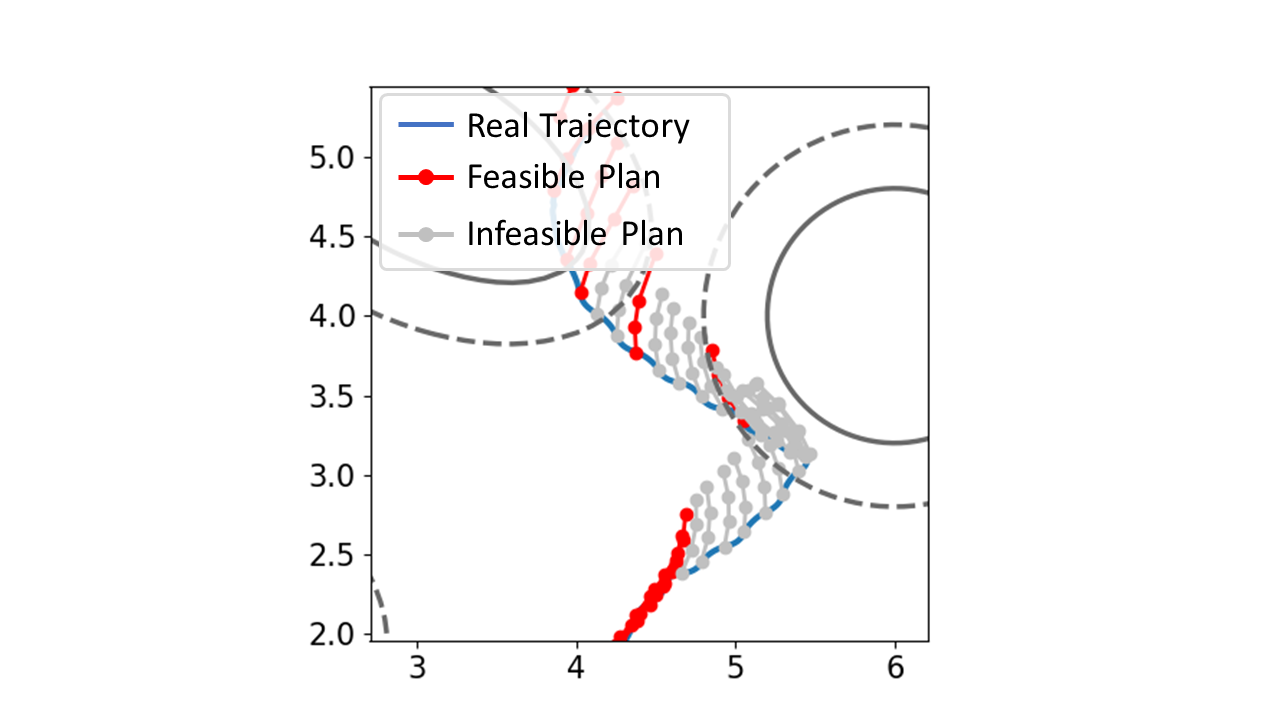}
         \caption{\small{DD-MPC}}
         \label{fig:cir_det_DD}
     \end{subfigure}
\caption{Planned and actual paths of both LIP-MPC and DD-MPC planners. The dashed line around obstacles is the extended safety region, which is used to form the CBF constraints. The infeasibility often occurs when the MPC planners violate the CBF constraints.}
\label{fig:env_sim_det}
\end{figure}

\figref{fig:env_sim_det} highlights the planned paths and actual paths of two planners at every step around the area when the failure occurs. \figref{fig:cir_det_LIP} shows the actual CoM trajectory of the robot in simulation closely following the planned paths from the LIP-MPC planner, thereby significantly improving the safety of the robot in navigation. Although some instances of infeasibility due to CBF constraint violations occurred with the LIP-MPC, the robot's CoM trajectories remained at a safe distance from obstacles, enabling collision-free navigation through real-time reactive planning.
In contrast, the significant model mismatch between the differential-drive model and the dynamics of bipedal robots led to discrepancies between DD-MPC's planned trajectories and the robot's actual CoM trajectories, as shown in \figref{fig:cir_det_DD}. This mismatch resulted in repeated infeasible solutions from the DD-MPC, culminating in the robot breaching the safety margin, colliding with an obstacle, and subsequently falling.

\begin{figure}
\centering
\vspace{2mm}
    \includegraphics[trim={3cm 0cm 3cm 0cm},clip,width=1\columnwidth]{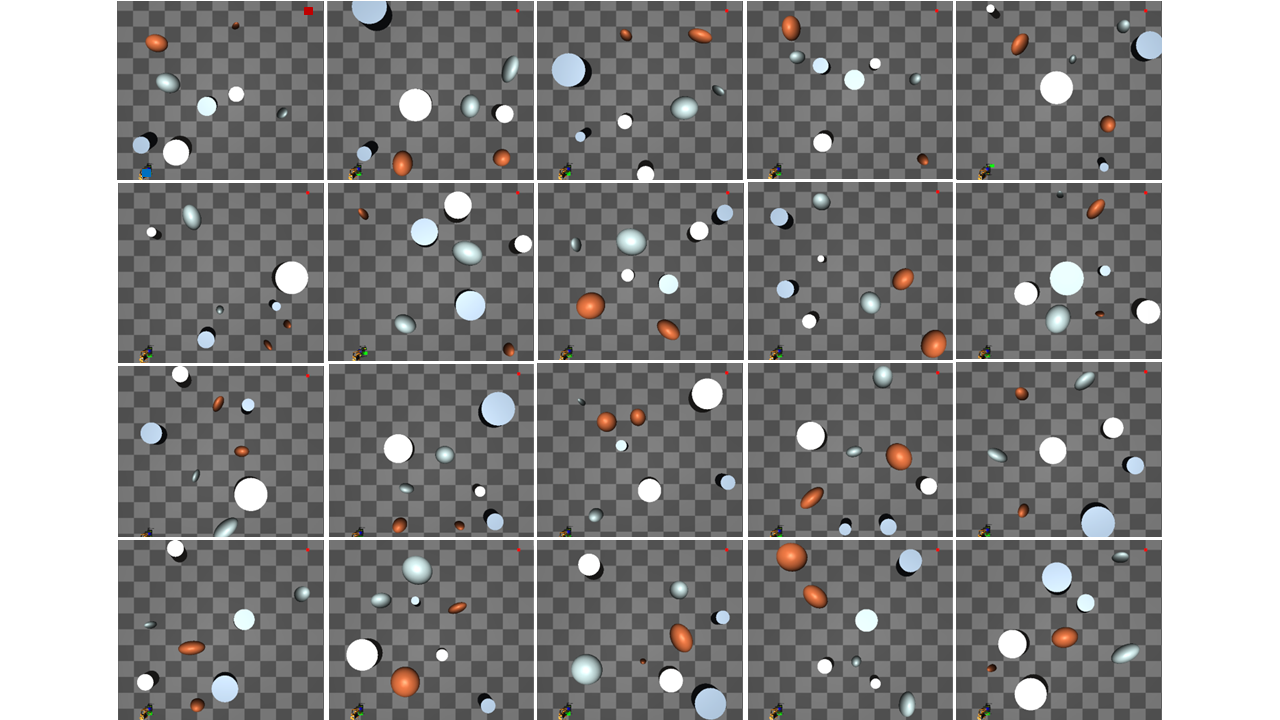}
\caption{Twenty distinctive environments populated with eight randomly sized and shaped obstacles.}
\label{fig:20_maps}
\end{figure}

\begin{table}[b]
\centering
\vspace{5mm}
\begin{tabular}{l l l l l } \hline
    Planner & Finish & Violate & Enter & Collide \\ \hline
    LIP-MPC & 20 & 15 & 10 & 0 \\ \hline
    DD-MPC & 4 & 20 & 17 & 16 \\ \hline
\end{tabular}
\caption{The results of applying the two planners to the 20 randomly generated scenarios.}
\label{table:1}
\end{table}

In an effort to rigorously compare the two planning approaches, we conducted simulations in 20 distinct environments, each populated with eight randomly sized and shaped obstacles, as illustrated in \figref{fig:20_maps}. Across all scenarios, the robot was tasked to navigate from a starting point at $[0,0]$ m to a goal point at $[10,10]$ m. The outcomes of these simulations are summarized in \tabref{table:1}. The results reveal that the LIP-MPC consistently led the robot to successfully navigate to its destination without colliding with any obstacles in all trials. Despite this, there were 15 instances of constraint violations, and in 10 of these cases, the robot briefly entered into the safety margins surrounding obstacles. Nonetheless, these violations did not result in the robot falling, likely because the violations were minor. Fine-tuning of control parameters, such as adjusting $\gamma$ to smaller values, could yield more cautious planning, reducing the robot's proximity to obstacles. However, this might also increase the chances of infeasibility in scenarios where navigable space is limited. Adjustments to kinematic constraints may also enhance optimization convergence, though infeasibility might still arise under particularly challenging conditions, such as in densely obstacle-laden environments. Moreover, the model mismatch between the LIP model and the actual robot would also make the optimization problem infeasible. 

On the other hand, the DD-MPC managed to complete the navigation task in only 4 out of the 20 trials. The remaining 16 attempts ended in collisions with obstacles, leading to the robot's fall. Constraint violations were noted in every test, with the robot entering the safety margins in 17 instances. Although similar adjustments could theoretically improve DD-MPC's performance, the fundamental mismatch between the differential-drive model and the actual dynamics of bipedal robots greatly compromises the safety and effectiveness of this planner.

%% file: sections/conclusions.tex
\section{Conclusion} 
\label{sec:conclusion}

In this paper, we presented a linear inverted pendulum (LIP) model-based Model Predictive Control (MPC) framework for unified path and gait planning of bipedal locomotion. The results demonstrated that our framework significantly enhances the safety and stability of the Digit robot in various clustered environments.  A key to our approach is the facilitation of smoother turning and dynamic speed adjustments during turns, which collectively bolster the system's reliability. Compared to traditional hierarchical planning methods, our model consistently delivered superior performance in all evaluated scenarios.
Despite its effectiveness, there is room for further refinement to achieve seamless real-time navigation. Future directions for this work include optimizing the structure of the optimization problem to enhance computational efficiency. Additionally, to tackle the challenge of infeasible planning scenarios, we intend to develop more sophisticated safety constraints and consider the incorporation of regularization techniques within the optimization problem. This adjustment aims to reduce the likelihood of constraint violations. Furthermore, exploring new models to fully harness the locomotion capabilities of bipedal robots remains a priority. Through these enhancements, we aspire to push the boundaries of bipedal robot navigation and operational agility.